\pdfoutput=1

\documentclass[11pt]{article}

\usepackage[preprint]{acl}

\usepackage{times}
\usepackage{latexsym}

\usepackage[T1]{fontenc}

\usepackage[utf8]{inputenc}

\usepackage{microtype}

\usepackage{inconsolata}

\usepackage{graphicx}

\usepackage[colorinlistoftodos]{todonotes}

\usepackage{amsthm}
\newtheorem{definition}{Definition}

\usepackage{hyperref}
\usepackage{booktabs}
\usepackage{arydshln}  
\usepackage{changepage}
\usepackage{pgfplots}
\usepackage{subcaption}
\usepackage{algorithm}
\usepackage{algpseudocode}
\usepackage{algorithmicx}
\usepackage{amsmath}
\usepackage{amssymb}
\usepackage{cleveref}
\usepackage[percent]{overpic}
\usepackage{xcolor}  
\usepackage{enumitem}
\usepackage[utf8]{inputenc}
\usepackage[T1]{fontenc}

\newcommand{\ourmodel}{\textsc{Attrieval}}

\title{Attention Reveals More Than Tokens: Training-Free Long-Context Reasoning with Attention-guided Retrieval}

\author{Yuwei Zhang, Jayanth Srinivasa$^2$, Gaowen Liu$^2$, Jingbo Shang$^1$\thanks{$\ $  Corresponding authors.} \\
University of California, San Diego$^1$ \quad Cisco$^2$ \\
  \texttt{\{yuz163, jshang\}@ucsd.edu}\\
  \texttt{\{jasriniv, gaoliu\}@cisco.com}
  }

\begin{document}
\maketitle
\begin{abstract}
Large Language Models (LLMs) often exhibit substantially shorter effective context lengths than their claimed capacities, especially when handling complex reasoning tasks that require integrating information from multiple parts of a long context and performing multi-step reasoning. Although Chain-of-Thought (CoT) prompting has shown promise in reducing task complexity, our empirical analysis reveals that it does not fully resolve this limitation. Through controlled experiments, we identify poor recall of implicit facts as the primary cause of failure, which significantly hampers reasoning performance. Interestingly, we observe that the internal attention weights from the generated CoT tokens can effectively ground implicit facts, even when these facts are not explicitly recalled. Building on this insight, we propose a novel training-free algorithm, {\ourmodel}, which leverages attention weights to retrieve relevant facts from the long context and incorporates them into the reasoning process. Additionally, we find that selecting context tokens from CoT tokens further improves performance. Our results demonstrate that {\ourmodel} enhances long-context reasoning capability notably on both synthetic and real-world QA datasets with various models.
\end{abstract}

\section{Introduction}

Recent advancements in long-context language models have unlocked the ability to process much larger input sequences~\cite{zaheer2020big,gu2023mamba,peng2023yarn,chen2023extending,chen2023longlora,jin2024llm,wang2024beyond}, achieving near perfect recall on retrieval tasks such as \emph{needle-in-a-haystack}~\cite{LLMTest_NeedleInAHaystack}. However, real-world applications—including multi-hop question answering~\cite{yang2018hotpotqa,trivedi2022musique}, document-level reasoning~\cite{mou2021narrative,dasigi2021dataset}, and multi-turn conversational agents~\cite{wu2024longmemeval} demand more than verbatim fact extraction, sometimes requiring aggregating information from scattered evidence into coherent conclusions. 
While existing models excel at locating explicit statements, their performance degrades significantly as context length increases for tasks requiring reasoning, even when all necessary facts are present in the input~\cite{hsieh2024ruler,kuratov2024babilong,ling2025longreason,bai2024longbench,zhang-etal-2024-bench}.
This discrepancy reveals a critical gap: strong single-hop retrieval capabilities do not inherently enable robust reasoning.

\begin{figure}
    \centering
    \includegraphics[width=0.99\linewidth]{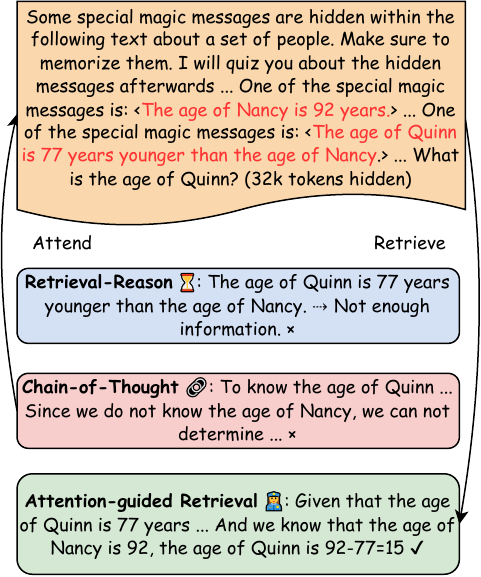}
    \caption{Both Retrieval-Reason (agentic framework) and Chain-of-Though (CoT) might suffer from poor recall of implicit facts. Our proposed {\ourmodel} leverage the internal attention weights to resolve this issue.}
    \label{fig:illustrative}
    \vspace{-5mm}
\end{figure}

Chain-of-Thought (CoT) reasoning~\cite{wei2022chain} offers a promising framework for complex tasks by decomposing reasoning into retrieval and inference steps that receives few attention in previous benchmarking.
The step-by-step CoT reasoning turns multi-hop questions into single-hop retrieval tasks that are easier to be solved by long-context models.
Yet, we observe that even with CoT, performance degrades sharply as context length increases. We hypothesize that this stems from failures to retrieve implicit facts—information critical for reasoning but lacking explicit surface cues as illustrated by \autoref{fig:illustrative}. To test this, we introduce \textbf{Deduction}, a diagnostic benchmark requiring models to (1) retrieve numerical facts from long contexts and (2) perform arithmetic reasoning. By analyzing responses for both fact recall and final accuracy, we find that the performance is mostly bottlenecked by the missed implicit (or second-hop) facts, not faulty arithmetic.

\begin{figure*}[ht!]
    \centering
    \begin{subfigure}[b]{0.49\textwidth}
        \centering
        \includegraphics[width=\textwidth, trim=0 41 0 0, clip]{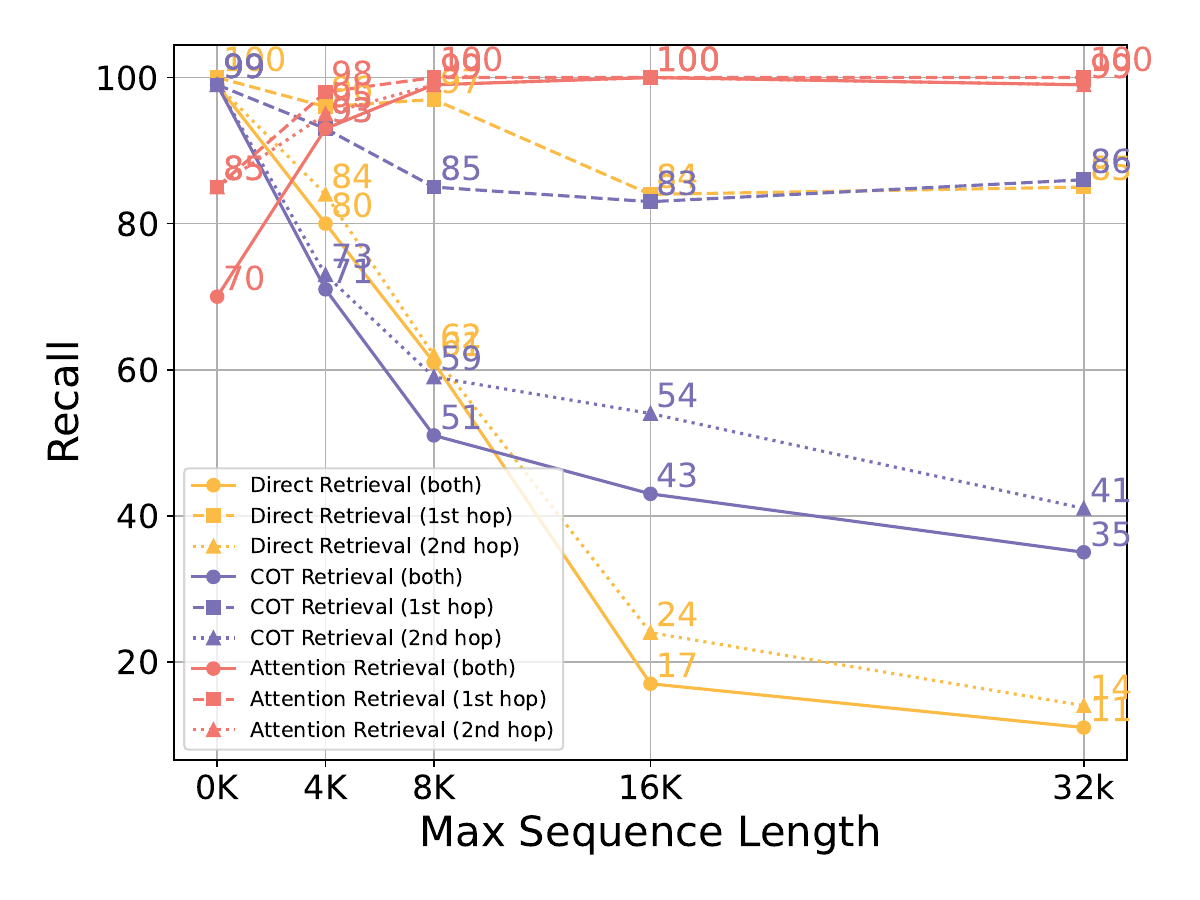}
        \caption{}
        \label{fig:recall_analysis}
    \end{subfigure}
    \hfill
    \begin{subfigure}[b]{0.49\textwidth}
        \centering
        \includegraphics[width=\textwidth, trim=0 41 0 0, clip]{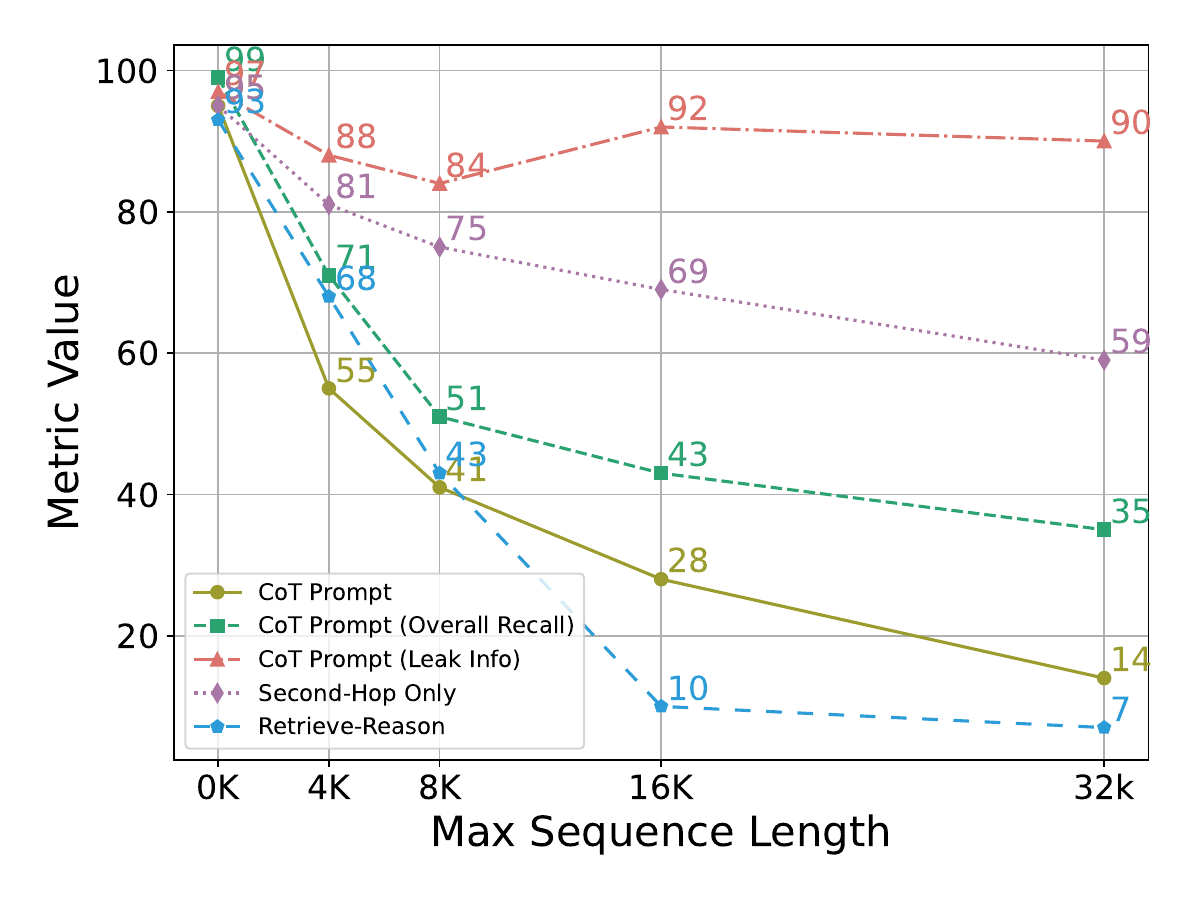}
        \caption{}
        \label{fig:retrieval_or_reasoning}
    \end{subfigure}
    \vspace{-3mm}
    \caption{Analysis on CoT tokens, including: (a) recall with various retrieval methods; (b) accuracy with various prompts and questions. See \autoref{sec:observation} for more details.}
    \label{fig:cot_analysis}
    \vspace{-5mm}
\end{figure*}

Notably, agentic frameworks have been explored to improve CoT in the literature by explicitly prompt the LLMs to retrieve-then-reason. For instance, \citet{zhang2024chain} proposed a multi-agent framework that distributes the long-context across multiple agents and then aggregates information through model collaboration. \citet{zhang2024steering} proposed an automatic attention steering framework that utilizes prompt-based method to elicit the model to generate useful facts and ``steer'' the attention weights. \citet{chen2023walking} proposed memory maze that summarizes the long-context into a hierarchical structure and then perform tree search during inference time. However, neither of them solve the implicit fact retrieval problem inherently (\autoref{fig:recall_analysis}) and might introduce laborious prompt engineering efforts or rely on strong close-source LLMs. Furthermore, CoT reasoning inherently outperforms agentic workflows by leveraging LLM's native generation of coherent, self-contained reasoning paths while maintaining computational efficiency and scalability.

In this paper, we first make the observation that the internal attention weights often highlight the overlooked implicit facts, suggesting a disconnect between latent retrieval signals (attention) and explicit generation.
Inspired by these findings, we propose \underline{Att}ention-guided \underline{Retrieval} (\ourmodel), a training-free framework that enhances long-context reasoning without compromising short-context performance or requiring laborious prompt engineering. {\ourmodel} operates in three key stages:
(1) The input context is partitioned into discrete facts, which are then ranked by their attention weights from intermediate CoT tokens.
(2) To counter the dominance of ``attention sink'' tokens, we filter out facts that appear in the top-$k$ attended positions for an excessive proportion of CoT tokens.
(3) We introduce a cross-evaluation framework to identify retriever tokens from the generated CoT sequence by measuring the KL-divergence between model predictions with and without the context.
The final retrieved facts are reintegrated into the context, enabling the model to reason over both explicit and previously overlooked implicit information.

Our works makes the following contributions:
\begin{itemize}[nosep,leftmargin=*]
    \item We introduce Deduction, a controlled benchmark for long-context reasoning, and identify retrieval failures—particularly for latent facts—as the primary bottleneck in existing methods.
    \item We demonstrate that attention weights encode latent factual relevance even when generated tokens fail to reference them explicitly, challenging the assumption that token outputs fully reflect model ``knowledge''.
    \item {\ourmodel} provides the first training-free solution that leverages attention patterns to bridge the gap between retrieval and reasoning, achieving state-of-the-art performance across both synthetic and realistic QA benchmarks (e.g., +47\% accuracy on Deduction and +11\% accuracy on MuSiQue on 32K context length).
\end{itemize}

\section{Preliminary}

We formally define \textit{long-context reasoning} task in this section.
\begin{definition}[Long-Context Reasoning]
Let $Q$ be a question (e.g., a natural-language query), and let
$\hat{I} = \{i_1, i_2, \dots, i_r\}$
be a set of \emph{informative} (or \emph{relevant}) facts needed to correctly answer $Q$. 
Let
$\hat{N} = \{n_1, n_2, \dots, n_s\}$
be a set of \emph{noisy} (or \emph{irrelevant}) facts. 
Define the \emph{long context} $C$ as the union of these two sets:
$C \;=\; \hat{I} \;\cup\; \hat{N}.$
Suppose there is an (ideal) \emph{reasoning function}
$R : \mathcal{Q} \times \mathcal{I} \;\to\; \mathcal{A},$
where $\mathcal{Q}$ is the space of all possible questions, $\mathcal{I}$ is the space of all possible informative-fact sets, and $\mathcal{A}$ is the space of all possible answers.

The \emph{long-context reasoning problem} is to construct a function
$\hat{R} : \mathcal{Q} \times \mathcal{C} \;\to\; \mathcal{A}$
that approximates $R$ when presented with the full long context $C$, i.e.,
$\hat{R}(Q,\, C) \;\approx\; R(Q,\, \hat{I}).$
\end{definition}

From a probabilistic point of view, long-context reasoning requires the model to be able to ``filter out'' (or marginalize) the noise $\hat{N}$ in posterior distribution:
\begin{equation}
    P(A=a|Q,\ C) \;\approx\; P(A=a|Q,\ \hat{I})
\end{equation}

\section{Analysis on CoT Tokens}\label{sec:observation}
A natural strategy for improving long-context reasoning is Chain-of-Thought (CoT) prompting~\cite{wei2022chain}, which enables models to strategically search through extended contexts~\cite{yu2023chain,li2024alr,li2024making}. The generated reasoning chain can decompose long-context reasoning into two subtasks: \emph{retrieval} and \emph{reasoning}. This process can be formulated as follows:
\begin{equation}
    P(A, Y|Q,\ C) = \underbrace{P(Y|Q,\ C)}_{\emph{retrieval}} \underbrace{P(A|Y,\ Q,\ C)}_{\emph{reasoning}}
\end{equation}
where $Y$ represents retrieved facts during \emph{retrieval} phase. While models can dynamically alternate between retrieval and reasoning to iteratively refine outputs, our experiments in \autoref{fig:retrieval_or_reasoning} demonstrate that CoT alone fails to mitigate the performance degradation in long-context scenarios. We identify two distinct failure modes: (1) search errors, where retrieved facts \(Y\) are incomplete or misaligned with \(Q\); or (2) reasoning errors, where the model misapplies logical rules despite accurate retrieval.
To dissect these issues, we first quantify the relative impact of each error type through empirical analysis. We then reveal a critical insight: transformers’ internal attention mechanisms exhibit stronger grounding to contextually relevant facts compared to explicit CoT-generated retrieval tokens. This finding suggests inherent limitations in relying solely on CoT’s discrete search phase for long-context understanding.

\subsection{Which is the Devil? Search or Reasoning?}
To systematically diagnose the interplay between search and reasoning errors, we require a benchmark where both retrieval validity (whether all necessary facts are recalled) and reasoning validity (whether logic is correctly applied) can be unambiguously evaluated. Existing long-context datasets often conflate these two aspects, as their open-ended questions and implicit grounding in context make it difficult to isolate failure modes.

To address this, we introduce \textbf{Deduction}, a diagnostic benchmark featuring synthetic reasoning tasks with explicit ground-truth retrieval requirements. Each task embeds a set of atomic facts (\emph{e.g.}, ``Nancy's age is 92'') within a long, distractor-filled context, followed by a deterministic question (\emph{e.g.}, ``What is Quinn's age?'') solvable only by recalling all relevant facts (\emph{e.g.}, ``Quinn is 77 years younger than Nancy'') and applying basic arithmetic. Crucially, our design ensures both controlled retrieval evaluation for both explicit and implicit facts and deterministic reasoning. See \autoref{sec:deduction} for details about dataset creation.


\noindent \textbf{Observations}
As illustrated in \autoref{fig:cot_analysis}, four key patterns emerge:
(1) First-hop recall (retrieving explicit facts like ``Nancy's age is 92'') remains robust (80–90\% across 4K–32K contexts), while second-hop recall (implicit dependencies like “Quinn’s age depends on Nancy”) drops sharply as sequence length increases, narrowing the gap between overall recall and second-hop recall (\autoref{fig:recall_analysis}).
(2) Despite being widely employed in agentic workflows, directly prompting the model retrieve useful information amplify the incomplete retrieve issue compared with a more natural CoT prompt (\autoref{fig:recall_analysis}).
(3) Final answer accuracy lags behind recall by 15–20\% (\autoref{fig:retrieval_or_reasoning}), indicating that even when models retrieve partial facts, they might fail to synthesize them into correct answers.
(3) Retrieval is the primary bottleneck. When explicitly prompted for second-hop facts (``What is Nancy's age?''), retrieval success improves by 35\% (\autoref{fig:retrieval_or_reasoning}, \textcolor[HTML]{A977A6}{Second-Hop Only}), confirming that models can reason accurately if retrieval is guaranteed.
(4) Appending ground-truth facts post-context (\textcolor[HTML]{F1766D}{Leak Info}) restores 85\% of 0K baseline performance (\autoref{fig:retrieval_or_reasoning}), yet a residual 15\% accuracy gap persists, likely due to attention dispersion over long sequences.
These results underscore that while reasoning errors occur, the dominant failure mode is retrieval: models struggle to retrieve implicit, interdependent facts from long contexts. The compounding effect of partial retrieval and flawed logic explains the steep performance decline in multi-hop tasks.

\subsection{Can Attention Weights Retrieve Latent Facts?}\label{sec:attention_retrieve}

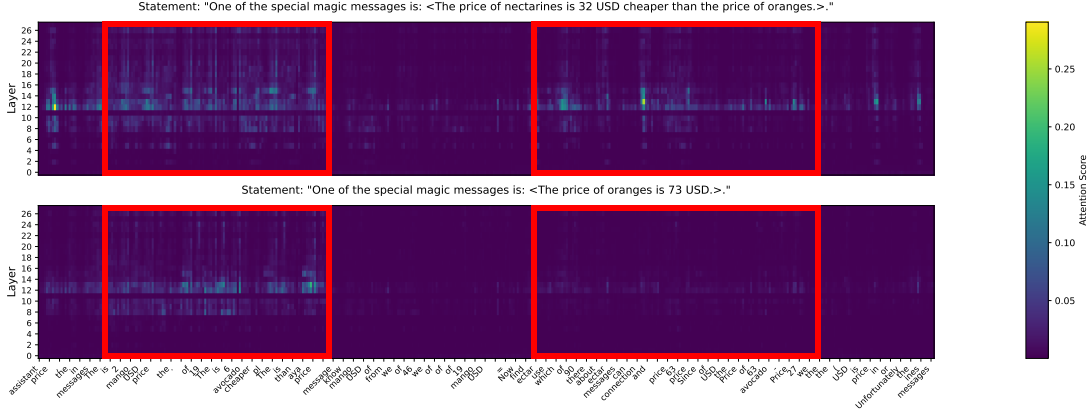
\begin{figure*}[t]
    \centering
    \begin{overpic}[width=0.90\linewidth, trim=0 20 0 0, clip]{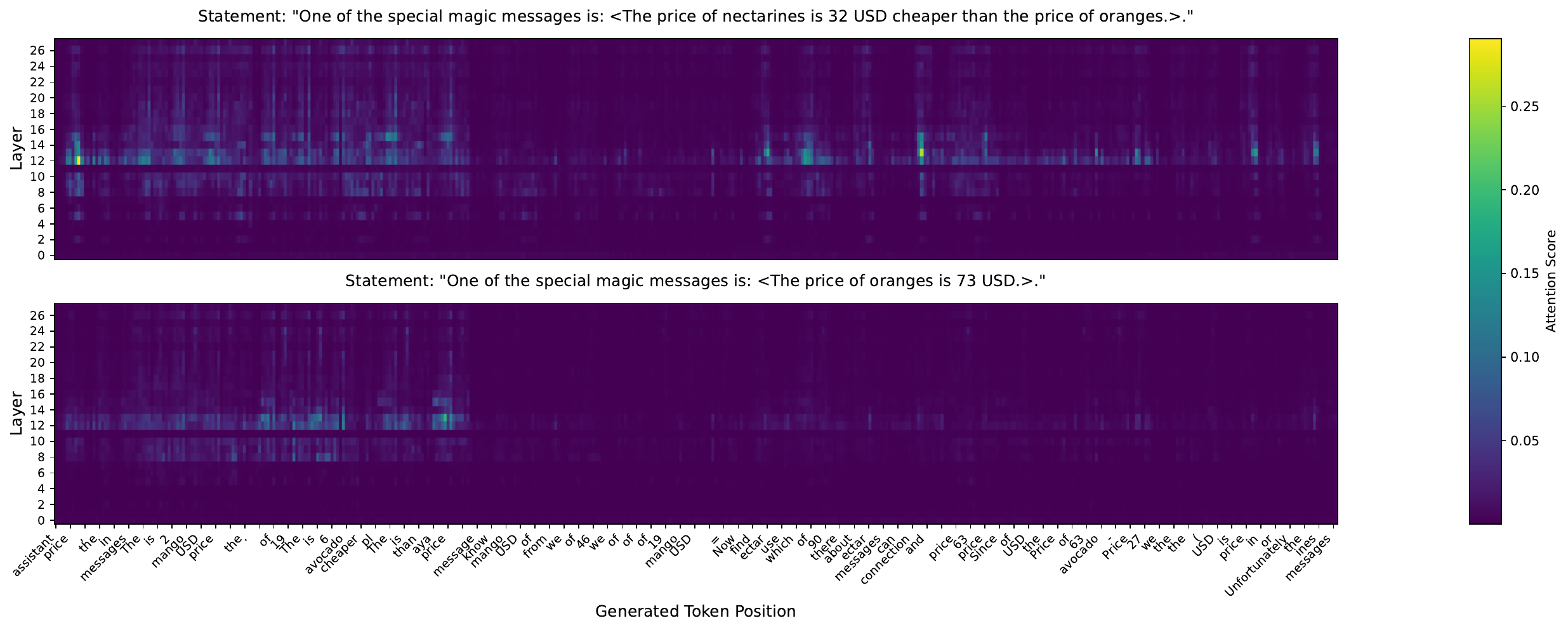}
        \put(9.8,5.5){\color{red}\linethickness{2pt}\framebox(20,13.2){}}
        \put(9.8,22.4){\color{red}\linethickness{2pt}\framebox(20,13.2){}}
        \put(49.0,22.4){\color{red}\linethickness{2pt}\framebox(25.5,13.2){}}
        \put(49.0,5.5){\color{red}\linethickness{2pt}\framebox(25.5,13.2){}}
    \end{overpic}
    \vspace{-3mm}
    \caption{Proportion of attention from generated tokens to the input prompt across layers.}
    \label{fig:heatmap}
    \vspace{-3mm}
\end{figure*}

While CoT prompting struggles to surface implicit facts through its explicit token generations, we find that the model's internal attention patterns reveal richer evidence of factual grounding. We hypothesize that this discrepancy arises because generated tokens represent high-level discretizations of latent states, potentially obscuring the model's sensitivity to specific input features. By contrast, attention weights provide continuous-valued signals that better preserve these fine-grained associations. This observation motivates our central investigation: \emph{Does the model internally attend to factual evidence that remains implicit in its generations?}
Through quantitative analysis of attention patterns (\autoref{fig:heatmap}), we demonstrate that the model allocates substantial attention to second-hop factual relationships, even when these fail to surface in CoT generations.
To formalize this analysis, let $t \in \{1,2,\dots,T\}$
denote the positions of the generated tokens and 
$i \in \{1,2,\dots,N\}$
denote the positions of the input tokens. For a given layer \( l \), we normalize the attention weights \( A^{(l)}_{t,i} \) so that they satisfy
$\sum_{i=1}^N A^{(l)}_{t,i} = 1.$
For a statement spanning input tokens indexed by 
$I_{\text{stmt}} \subseteq \{1,2,\dots,N\}$,
we compute the aggregated attention score for each layer and generated token as
\begin{equation}
    H_{\text{stmt}}(l,t)=\sum_{i\in I_{\text{stmt}}} A^{(l)}_{t,i}.
\end{equation}
Our case study in \autoref{fig:heatmap} reveals two key patterns: (1) Early generated tokens exhibit heightened attention to both first-hop and second-hop factual statements (red boxes), despite the CoT ultimately failing to verbalize the latter, and (2) While first-hop attention resurfaces in later tokens, second-hop attention remains suppressed.
We further show in \autoref{fig:ranking} that the rankings of attention weights spent on the statement tokens are usually high.
Notably, second-hop statements achieve comparable ranking positions to first-hop statements during initial generated tokens. Nonetheless, it remains challenging to extract these statements from the overall input, as high attention weights are also assigned to other irrelevant tokens, such as those at the beginning of the prompt and the most recent tokens~\cite{xiao2023efficient,han2023lm}.

\section{Methodology}

\definecolor{mygreen}{rgb}{0.0, 0.5, 0.0}  
\newcommand{\inc}[1]{\textcolor{mygreen}{#1}}
\newcommand{\dec}[1]{\textcolor{red}{#1}}    
\setlength{\tabcolsep}{12pt} 
\renewcommand{\arraystretch}{0.90} 
\begin{table*}[t]
\centering
\small
\setlength\dashlinedash{0.5pt}
\setlength\dashlinegap{1.5pt}
\scalebox{0.98}{
\begin{tabular}{llcccccc}
\hline\hline
Model & Method & 0K & 4K & 8K & 16K & 32K & Overall \\
\hline\hline
\multicolumn{8}{c}{Deduction} \\
\hline\hline
& CoT                   & 95  & 55  & 41  & 28  & 14  & 47 \\
Llama-3.2-3B-Instruct  & \ourmodel    & \inc{97}  & \inc{76}  & \inc{75}  & \inc{63}  & \inc{57}  & \inc{74} \\
& \ourmodel-kl         & \inc{96}  & \inc{79}  & \inc{80}  & \inc{77}  & \inc{61}  & \inc{79} \\
\hdashline
& CoT                   & 99  & 96  & 93  & 74  & 70  & 86 \\
Llama-3.1-8B-Instruct  & \ourmodel    & 99 & \inc{97} & \inc{99} & \inc{92} & \inc{81} & \inc{94} \\
& \ourmodel-kl         & \inc{100} & \inc{98} & \inc{96} & \inc{91} & \inc{83} & \inc{94} \\
\hdashline
& CoT                   & 96  & 46  & 50  & 34  & 28  & 51 \\
Qwen2.5-3B-Instruct    & \ourmodel    & \inc{99}  & \inc{59}  & \inc{60}  & \inc{56}  & \inc{40}  & \inc{63} \\
& \ourmodel-kl         & \inc{99}  & \inc{55}  & \inc{55}  & \inc{48}  & \inc{56}  & \inc{63} \\
\hline\hline
\multicolumn{8}{c}{MuSiQue} \\
\hline\hline
& CoT                   & 78  & -   & 25  & 24  & 18  & 36 \\
Llama-3.2-3B-Instruct  & \ourmodel    & \dec{71}  & -   & \inc{38}  & \inc{33}  & \inc{23}  & \inc{41} \\
& \ourmodel-kl         & \dec{74}  & -   & \inc{43}  & \inc{33}  & \inc{29}  & \inc{45} \\
\hdashline
& CoT                   & 78  & -   & 61  & 45  & 29  & 53 \\
Llama-3.1-8B-Instruct  & \ourmodel    & \inc{87}  & -   & \inc{64}  & \inc{55}  & \inc{45}  & \inc{63} \\
& \ourmodel-kl         & \inc{79}  & -   & \inc{68}  & \inc{58}  & \inc{41}  & \inc{62} \\
\hdashline
& CoT                   & 69  & -   & 42  & 33  & 23  & 42 \\
Qwen2.5-3B-Instruct    & \ourmodel    & \inc{72}  & -   & \inc{48}  & \inc{39}  & \dec{18}  & \inc{44} \\
& \ourmodel-kl         & \dec{64}  & -   & \inc{45}  & \inc{36}  & \dec{22}  & \dec{42} \\
\hline\hline
\multicolumn{8}{c}{HotpotQA} \\
\hline\hline
& CoT                   & 71  & 68  & 62  & 57  & 58  & 63 \\
Llama-3.2-3B-Instruct  & \ourmodel    & \inc{73} & \dec{58} & \inc{66} & \inc{61} & \inc{59} & 63 \\
& \ourmodel-kl         & \dec{68} & \dec{67} & \inc{66} & \inc{61} & \dec{55} & 63 \\
\hdashline
& CoT                   & 71  & 71  & 71  & 70  & 68  & 70 \\
Llama-3.1-8B-Instruct  & \ourmodel    & \inc{74} & \inc{72} & \dec{69} & \inc{72} & \inc{69} & \inc{71} \\
& \ourmodel-kl         & \dec{69} & \dec{71} & \dec{70} & \dec{70} & \dec{67} & \dec{69} \\
\hdashline
& CoT                   & 64  & 61  & 60  & 54  & 54  & 59 \\
Qwen2.5-3B-Instruct    & \ourmodel    & \inc{66} & \dec{58} & \dec{52} & \inc{58} & \dec{48} & \dec{56} \\
& \ourmodel-kl         & \dec{63} & \dec{60} & \dec{58} & \inc{55} & \dec{50} & \dec{57} \\
\hline\hline
\end{tabular}
}
\caption{Main results with color annotations. Green numbers exceed CoT; red numbers are lower than CoT. For MuSiQue, the context already exceeds 4k tokens.}
\label{tab:main}
\end{table*}

\begin{algorithm}[t]
\caption{Attention-Guided Retrieval (\ourmodel)}
\label{alg:attret}
\begin{algorithmic}[1]
\Require Input context $\mathcal{X}$, generated CoT tokens $\{t_1, \dots, t_T\}$, layers $\mathcal{L}$, top-$k$ threshold, frequency threshold $\tau$, min tokens $m$, max facts $n$
\Ensure Retrieved facts $\mathcal{F}_{\text{retrieved}}$

\State \textbf{Stage 1: Multi-Layer Attention Aggregation}
\For{each generated token $t \in \{1, \dots, T\}$}
    \For{each input token $i \in \mathcal{X}$}
        \State Compute $\quad \bar{A}_{t,i}$ via \autoref{eq:attention_aggregate}
    \EndFor
\EndFor

\State \textbf{Stage 2: Common Facts Filtering}
\State Segment $\mathcal{X}$ into facts $\{c\}$ via punctuation
\For{each generated token $t$}
    \State Identify top-$k$ tokens: $\mathcal{T}_t$ via \autoref{eq:topk_token}
\EndFor
\For{each fact $c$}
    \State Compute frequency: $f(c)$ via \autoref{eq:frequency}
\EndFor
\State Filter sinks: $\mathcal{F}_{\text{filtered}} \gets \{c : f(c) < \tau\}$

\State \textbf{Stage 3: Fact Scoring \& Selection}
\For{each fact $c \in \mathcal{F}_{\text{filtered}}$}
    \State Aggregate fact score: $s(c)$ via \autoref{eq:final_score}
\EndFor
\State Sort facts by $s(c)$, filter length $\geq m$ tokens
\State Return $\mathcal{F}_{\text{retrieved}} \gets \text{top-}n$ facts

\end{algorithmic}
\end{algorithm}

Inspired by the previous observations that the attention weights perform better at grounding in the long-context setting, we now introduce a novel algorithm that improves long-context reasoning without any additional training or extensive prompt engineering. Intuitively, the proposed algorithm performs attention-based retrieval based on the generated CoT tokens, and then incorporate them for reasoning.

Formally, given a pre-defined set of layers $\cal L$, we first aggregate the attention over heads and layers,
\begin{equation}
    \bar{A}_{t,i} = \frac{1}{|\mathcal{L}|} \sum_{l \in \mathcal{L}} \left( \frac{1}{H} \sum_{h=1}^{H} A^{(l,h)}_{t,i} \right).
    \label{eq:attention_aggregate}
\end{equation}
The input sequence is segmented into discrete facts \( \{ c \} \) based on punctuations. Each input token $i$ is mapped to its corresponding fact $c(i)$. For each generated token $t$, we identify the top-$k$ input tokens with the highest aggregated attention scores, denoting their indices by the set \( \mathcal{T}_t \).
\begin{equation}
    \mathcal{T}_t = \arg\underset{i}{\text{top-}k}(\bar{A}_{t,i})
    \label{eq:topk_token}
\end{equation}
We then define the frequency of a fact \( c \) as
\begin{equation}
    f(c) = \frac{1}{T} \sum_{t=1}^T \mathbb{I}\left\{ c \in \{ c(i) : i \in \mathcal{T}_t \} \right\},
    \label{eq:frequency}
\end{equation}
where \( \mathbb{I}\{\cdot\} \) is the indicator function. Facts with \( f(c) \geq \tau \) (where $\tau$ is a threshold) are filtered as potential attention sinks~\cite{xiao2023efficient}---frequently attended tokens that provide little informational value. For remaining facts, we compute a relevance score by first averaging the aggregated attention over all generated tokens for each input token, and then averaging these scores over all tokens belonging to fact \( c \). Concretely, if \( I_c = \{ i : c(i) = c \} \), then the fact score is defined as
\begin{equation}
s(c) = \frac{1}{|I_c|} \sum_{i \in I_c} \left( \frac{1}{T} \sum_{t=1}^T \bar{A}_{t,i} \right).
\label{eq:final_score}
\end{equation}
These scores \( s(c) \) provide a measure of relevance between facts and generated tokens. We then take the top-$n$ facts while filtering out those with less than $m$ tokens. These facts are then incorporated into the context for generating the final answers. We can also select a subset of tokens to calculate final score as illustrated in the next paragraph. The prompt we used for this procedure requires minimal design. See \autoref{sec:prompt} for prompts and Algorithm~\ref{alg:attret} for algorithm procedure.

\noindent \textbf{Cross-Evaluation for Token Selection.}
As shown in \autoref{fig:heatmap}, we observe that there exist two kinds of tokens: \emph{retriever tokens} that cite the context and spread more attention on the ground truth facts; \emph{reasoner tokens} that focuses on reasoning with previously cited context. We hypothesize that the \emph{retriever tokens} might be better at retrieving relevant information from the context. Therefore, in this section, we propose a simple method to automatically detect \emph{retriever tokens} via cross-evaluation (shown in Algorithm~\ref{alg:token_selection}). Given context $C$ and question $Q$, the model evaluates the generate CoT tokens with both a long prompt ${\cal P}_L(C,\ Q)$ and a short prompt ${\cal P}_S(Q)$. The token-wise KL divergence $D_{KL}(P_L^{(t)}||P_S^{(t)})$ identifies tokens where contextual information most significantly alters their predictions. We then simply take the top-$s$ tokens as the selected \emph{retriever tokens} for \autoref{eq:final_score}.


\begin{algorithm}[t]
\caption{Cross-Evaluation Token Selection}
\label{alg:token_selection}
\begin{algorithmic}[1]
\Require Context $C$, question $Q$, token count $s$, Model $M$
\Ensure Selected retriever tokens $\mathcal{T}_{\text{retrieve}}$

\State Generate token distributions:
\State $\quad P_L^{(1:T)} \gets M(\mathcal{P}_L(C, Q))$ \Comment{Long prompt}
\State $\quad P_S^{(1:T)} \gets M(\mathcal{P}_S(Q))$ \Comment{Short prompt}

\State Compute token-wise divergence:
\For{each token $t \in \{1,\dots,T\}$}
    \State $D_{\text{KL}}^{(t)} \gets D_{\text{KL}}\left(P_L^{(t)} \parallel P_S^{(t)}\right)$
\EndFor

\State $\mathcal{T}_{\text{retriever}} \gets \arg\underset{t}{\mathrm{top\mbox{-}}s}(D_{\text{KL}}^{(t)})$

\end{algorithmic}
\end{algorithm}

\section{Main Results}\label{sec:results}

\begin{figure*}[t]
    \centering
    \begin{subfigure}[b]{0.32\textwidth}
        \centering
        \includegraphics[width=0.95\textwidth, trim=18.1 21 18 0, clip]{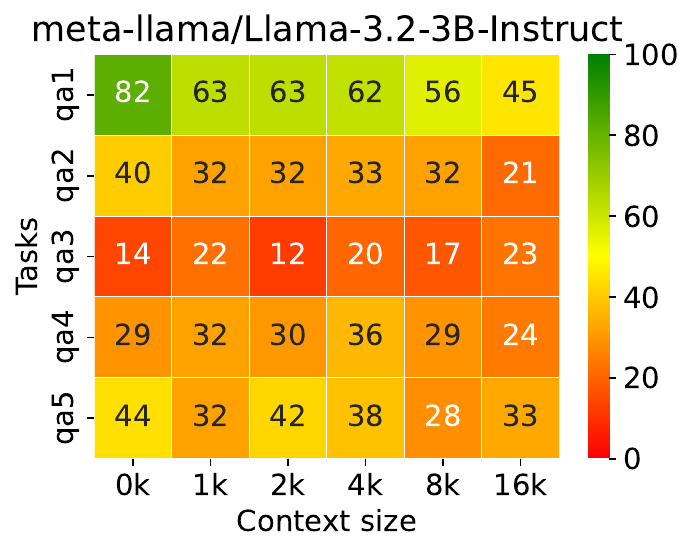}
    \end{subfigure}
    \begin{subfigure}[b]{0.32\textwidth}
        \centering
        \includegraphics[width=0.95\textwidth, trim=18.1 21 18 0, clip]{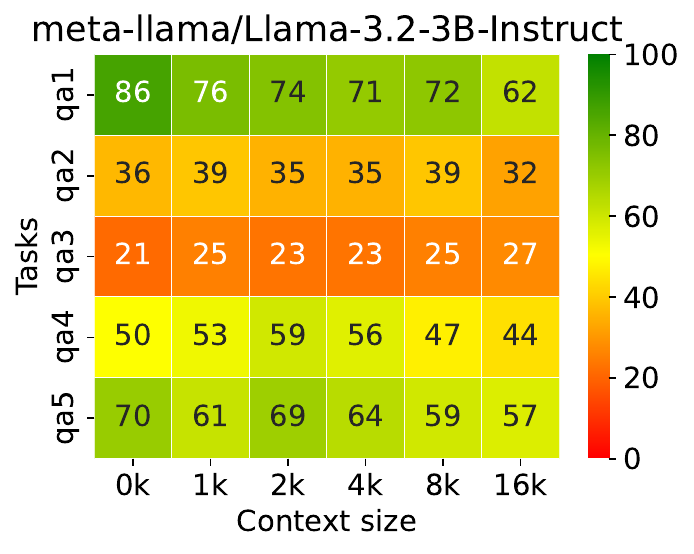}
    \end{subfigure}
    \hfill
    \begin{subfigure}[b]{0.32\textwidth}
        \centering
        \includegraphics[width=0.95\textwidth, trim=18.1 21 18 0, clip]{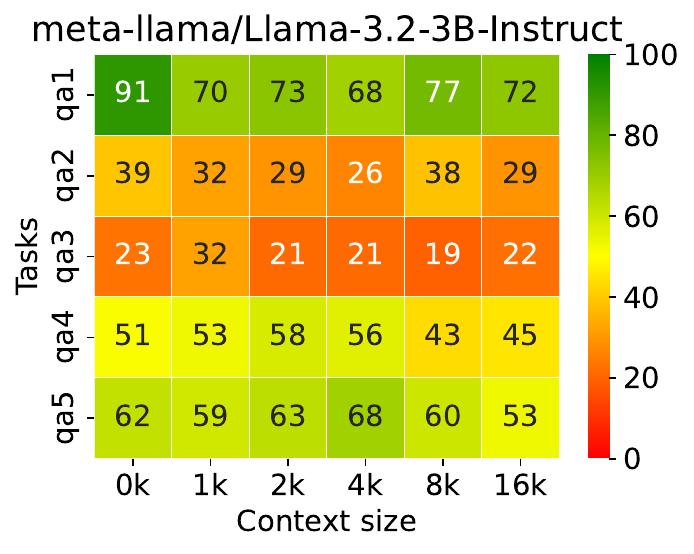}
    \end{subfigure}
    \\
    \begin{subfigure}[b]{0.32\textwidth}
        \centering
        \includegraphics[width=0.95\textwidth, trim=18.1 21 18 0, clip]{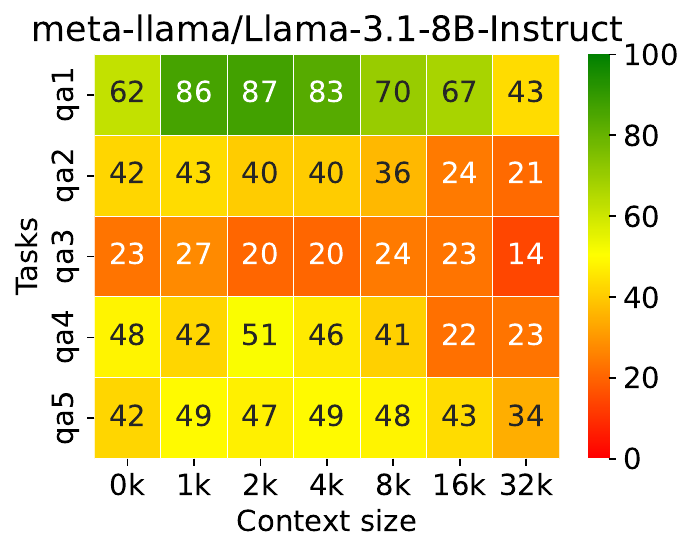}
    \end{subfigure}
    \hfill
    \begin{subfigure}[b]{0.32\textwidth}
        \centering
        \includegraphics[width=0.95\textwidth, trim=18.1 21 18 0, clip]{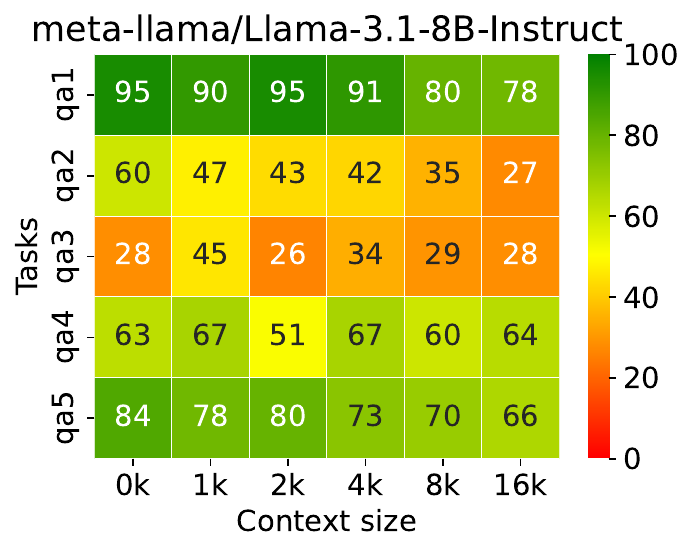}
    \end{subfigure}
    \hfill
    \begin{subfigure}[b]{0.32\textwidth}
        \centering
        \includegraphics[width=0.95\textwidth, trim=18.1 21 18 0, clip]{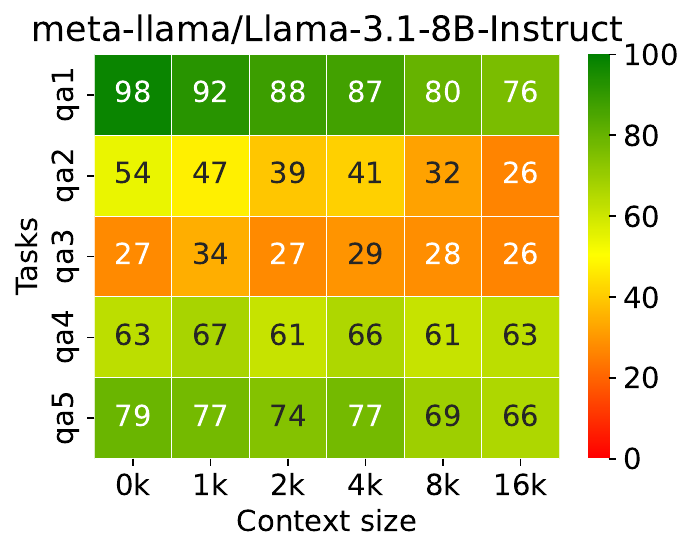}
    \end{subfigure}
    \\
    \begin{subfigure}[b]{0.32\textwidth}
        \centering
        \includegraphics[width=0.95\textwidth, trim=18.1 21 18 0, clip]{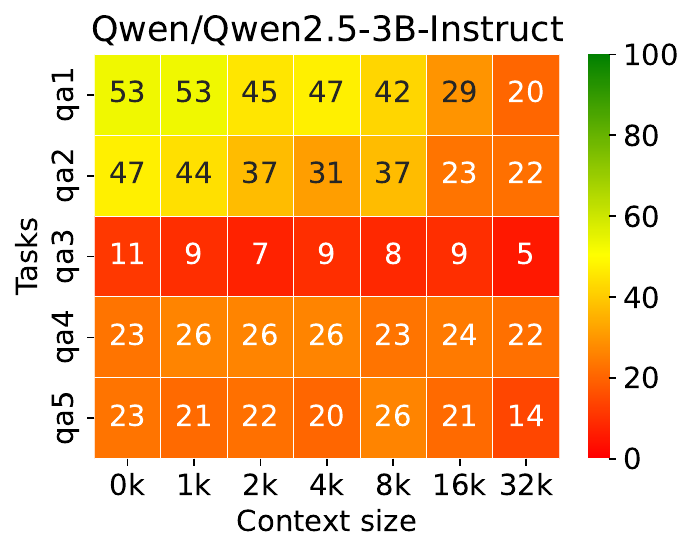}
        \caption{CoT}
    \end{subfigure}
    \hfill
    \begin{subfigure}[b]{0.32\textwidth}
        \centering
        \includegraphics[width=0.95\textwidth, trim=18.1 21 18 0, clip]{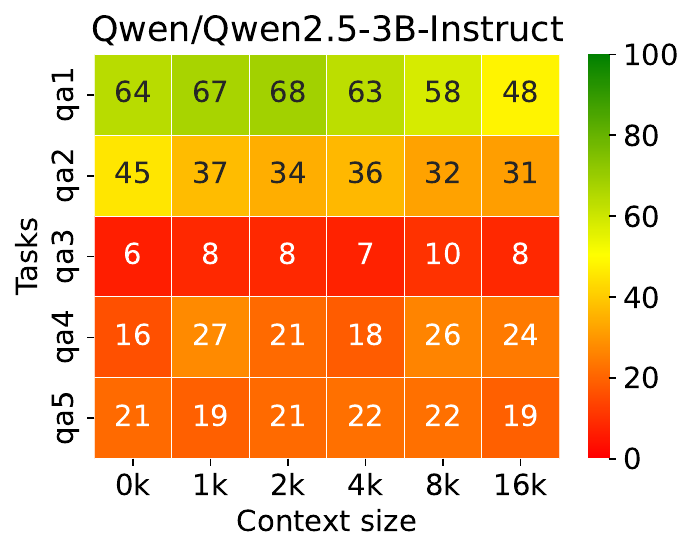}
        \caption{\ourmodel}
    \end{subfigure}
    \hfill
    \begin{subfigure}[b]{0.32\textwidth}
        \centering
        \includegraphics[width=0.95\textwidth, trim=18.1 21 18 0, clip]{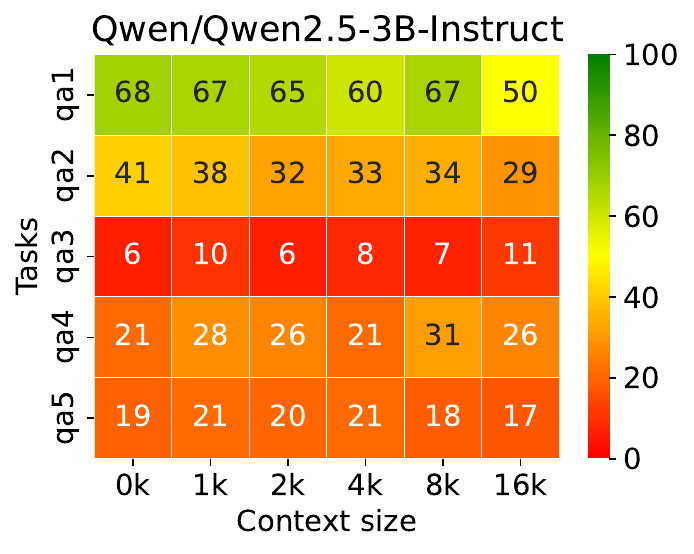}
        \caption{\ourmodel-kl}
    \end{subfigure}
    \caption{BABILONG results. Greener colors represent higher scores.}
    \label{fig:babilong}
    \vspace{-4mm}
\end{figure*}

\begin{table*}[t]
\centering
\small
\scalebox{0.98}{
\begin{tabular}{llccccccc}
\hline\hline
Tokens From & Select Strategy & 0K & 4K & 8K & 16K & 32K & Overall \\
\hline\hline
CoT    & None        & 95 & 55 & 41 & 28  & 14  & 47 \\
CoT    & All         & 97 & 76 & 75 & 63  & 57  & 74 \\
Random & All         & \textbf{99} & 61 & 60 & 55  & 48  & 65 \\
CoT    & First-s     & 98 & 66 & 56 & 49  & 39  & 62 \\
CoT    & Random-s    & 95 & 76 & 69 & 62  & 60  & 72 \\
CoT    & KL top-s    & 96 & \textbf{79} & \textbf{80} & \textbf{77}  & \textbf{61}  & \textbf{79} \\
\hline\hline
\end{tabular}
}
\caption{Analysis on the generated tokens used in {\ourmodel} to calculate attention matrices and the retriever token selection strategy. Studied dataset and model are Deduction and \texttt{meta-llama/Llama-3.2-3B-Instruct}.}
\label{tab:analysis}
\vspace{-5mm}
\end{table*}

\subsection{Experimental Setting}
In this paper, we mainly study three open-source models \texttt{meta-llama/Llama-3.2-3B-Instruct}, \texttt{meta-llama/Llama-3.1-8B-Instruct} and \texttt{Qwen/Qwen2.5-3B-Instruct}. In our preliminary experiments, we found that attention weights from higher layers can better ground to the context, we thus always choose the last $1/4$ layers as $\cal L$. We fix $k=50,\ \tau=0.99$ for filtering common facts, and minimum fact length $m=3$. We always choose $n=10$ facts for all the experiments. For token selection strategies, we consistently use $s=10$. All the experiments are done on a single A100 GPU.

\subsection{Evaluation Dataset}
We evaluate on both synthetic and realistic QA datasets. For synthetic QA, we evaluate with Deduction dataset with $2$ main entities and $6$ distraction entities. For realistic QA, we choose HotpotQA~\cite{yang2018hotpotqa} and MuSiQue~\cite{trivedi2022musique} since they mainly aim for multi-hop QA task. We follow RULER~\cite{hsieh2024ruler} for dataset creation. Furthermore, we evaluate on a more challenging benchmark dataset BABILONG~\cite{kuratov2024babilong} that requires the algorithm to be sensitive to the order of facts presented in the long context. For RULER datasets, we evaluate up to 32K context length, and 16K for BABILONG due to limited computational resource. We employ the same evaluation metric as proposed in each benchmark dataset. Due to limited computation resource, we evaluate $100$ examples for each of the length and dataset.

\subsection{Comparison on Benchmark Datasets}
Our experiments evaluate the performance of {\ourmodel} and its variant {\ourmodel}-kl across three open-source models and four datasets. The results summarized in \autoref{tab:main} and \autoref{fig:babilong}, demonstrate consistent improvements over the baseline CoT prompting, particularly in tasks requiring complex reasoning and long-context understanding. We conclude several key findings in the following, and show case studies in the Appendix:

\noindent \textbf{Superiority of {\ourmodel} over CoT}
On Deduction, {\ourmodel} and {\ourmodel}-kl significantly outperform CoT across all models. For instance, Llama-3.2-3B-Instruct with {\ourmodel}-kl achieves an overall score of 79 (vs. CoT's 47), while Qwen2.5-3B-Instruct improves from 51 (CoT) to 63 ({\ourmodel}). Gains are especially pronounced at longer context lengths (16K–32K), where {\ourmodel}-kl mitigates performance degradation (e.g., Llama-3.2-3B-Instruct at 32K: 61 \emph{vs.} CoT's 14).
On MuSiQue, {\ourmodel}-based methods exhibit stronger robustness. Llama-3.1-8B-Instruct with {\ourmodel} achieves an overall score of 63, surpassing its CoT counterpart by 21 points. Even smaller models like Qwen2.5-3B-Instruct show improvements ({\ourmodel}: 44 vs. CoT: 42).
On HotpotQA, {\ourmodel}-based methods perform modest. Llama-3.1-8B-Instruct with {\ourmodel} achieves a 71 overall score (\emph{vs.} CoT's 59), but smaller models like Qwen2.5-3B-Instruct show narrower margins ({\ourmodel}: 56 vs. CoT: 59).
We also notice that larger models (e.g., Llama-3.1-8B-Instruct) consistently outperform smaller counterparts when paired with {\ourmodel}, highlighting synergies between method efficacy and model capacity. For example, on MuSiQue, Llama-3.1-8B-Instruct with {\ourmodel} scores 63, far exceeding Qwen2.5-3B-Instruct's 44. We also show case studies in \autoref{sec:case_study}. And refer to \autoref{fig:recall_analysis} for the recall analysis of retrieved facts.

\noindent \textbf{Effectiveness of {\ourmodel}-kl Variant}
The {\ourmodel}-kl variant consistently matches or exceeds the base {\ourmodel} method. For example, on Deduction with Llama-3.2-3B-Instruct, {\ourmodel}-kl achieves a 79 overall score (\emph{vs.} {\ourmodel}'s 74), driven by superior performance at 16K (77 vs. 63). This suggests that integrating our proposed token selection strategy enhances retrieval performance.


\subsection{How does Token Selection Affect Performance?}

\setlength{\tabcolsep}{12pt}
\renewcommand{\arraystretch}{0.9} 

We study the effect of varying the generated tokens used for calculating attention scores and the retriever token selection strategy. Specifically, we first treat a paragraph of $150$ random words as if they are generated CoT tokens and proceed with {\ourmodel} algorithm. Surprisingly, as shown in \autoref{tab:analysis}, we found that even though the tokens are completely irrelevant with the context, they can still improve the performance over the vanilla CoT.
We further study the effect of token selection strategy. We propose several variants against our proposed KL-divergence based selection. ``First-s'' means we only select the first $s$ tokens in the sequence. ``Random-s'' means we select random $s$ tokens. From \autoref{tab:analysis}, we found that our proposed strategy performs the best among others. However, we do notice that on Qwen models, our strategy does not perform better than using all tokens, and we hypothesize that this is because Qwen models tend to generate longer CoT and selecting more tokens could help.

\section{Related Works}
\noindent \textbf{Long-context Reasoning}
Architectural innovations, such as modified positional encodings~\cite{chen2023extending,peng2023yarn,jin2024llm,chen2023longlora}, sparse attention mechanisms~\cite{zaheer2020big,lou2024sparser}, RNN-like models~\cite{gu2023mamba,peng2023rwkv} have enabled efficient processing of extended sequences while mitigating computational costs, as surveyed in~\citet{wang2024beyond}. However, challenges persist in multi-hop reasoning, where models exhibit sensitivity to noisy contexts~\cite{bai2024longbench,hsieh2024ruler,kuratov2024babilong,ling2025longreason,zhang-etal-2024-bench,wu2024longmemeval}.
To tackle this problem, recent research often employ fine-tuning based approaches that either focus on collecting complex long-context training data~\cite{li2024large,an2024make,chen2024long} or training the model to retrieve and cite the context before generating the answers~\cite{li2024alr,li2024making,yu2023chain}. While effective, these approaches face two key limitations: collecting high-quality long-context data is prohibitively expensive, and excessive specialization risks degrading performance on short-context tasks.
On the other hand, training-free agentic workflows are proposed improve long-context capability~\cite{zhang2024chain,chen2023walking,zhang2024steering}.
This work argues that these approaches does not inherently solve implicit fact retrieval problem.

\noindent \textbf{Attention-guided Retrieval}
Unlike traditional retrieval-augmented generation (RAG) pipelines that rigidly separate retrieval and generation stages, recent approaches leverage attention mechanisms to dynamically guide retrieval process. Notably, \citet{jiang2022retrieval} unifies retrieval and geenration in a single Transformer. \citet{jiang2023active,li2022contrastive} uses attention distribution guide or trigger retrieval. \citet{wu2022memorizing,borgeaud2022improving} introduce memory banks into Transformers via cross-attention. This work makes the observation that attention from CoT tokens can improve reasoning capability over long-context.

\section{Discussion and Conclusion}

This work starts by making several key observations on the Chain-of-Thought (CoT) of long-context reasoning tasks: (1) CoT struggles with multi-hop reasoning mainly due to incomplete retrieval of implicit facts; (2) attention patterns from intermediate CoT tokens consistently highlight relevant facts, even when those facts remain unmentioned in generated text.
We then present {\ourmodel}, a novel training-free framework that enhances long-context reasoning by grounding retrieval in the latent signals of transformer attention mechanisms. By identifying and reintegrating these retrieved facts, {\ourmodel} mitigates the performance degradation of LLMs on tasks requiring multi-hop reasoning over extended contexts. Our results on Deduction, BABILong, and real-world benchmarks like MuSiQue demonstrate its broad applicability and robustness. This work advances the understanding of how attention mechanisms can be harnessed to align context retrieval and reasoning, offering a lightweight yet effective solution.
Our work also spurs two potential future directions: (1) iteratively combine CoT generation and attention-guided retrieval based on the model uncertainty; (2) utilize attention weights from generated CoT as a supervision signal to better finetune long-context model.

\section*{Limitations}

Despite its effectiveness on various tasks and models, we point the following limitations of {\ourmodel}: (1) it requires two steps of response generation---one for acquiring attention matrix and the other for answer generation---which approximately doubles the inference costs. Future work could explore when to early stop the first-round generation and start retrieval. (2) {\ourmodel} can be effectively applied on applications with shorter CoT. However, when it is applied on long-form generation tasks, {\ourmodel} should be applied iteratively during the generation. (3) {\ourmodel} still does not completely solve long-context performance degradation. There is still a minor issue that the reasoning steps can be distracted by excessive attentions spreaded on previous sequence. Future work could explore context reduction guided by attention weights.

\section*{Ethics Consideration}
This paper only studies datasets in English language.

\bibliography{custom}

\appendix

\begin{figure*}
    \centering
    \includegraphics[width=0.95\textwidth, trim=0 0 0 20, clip]{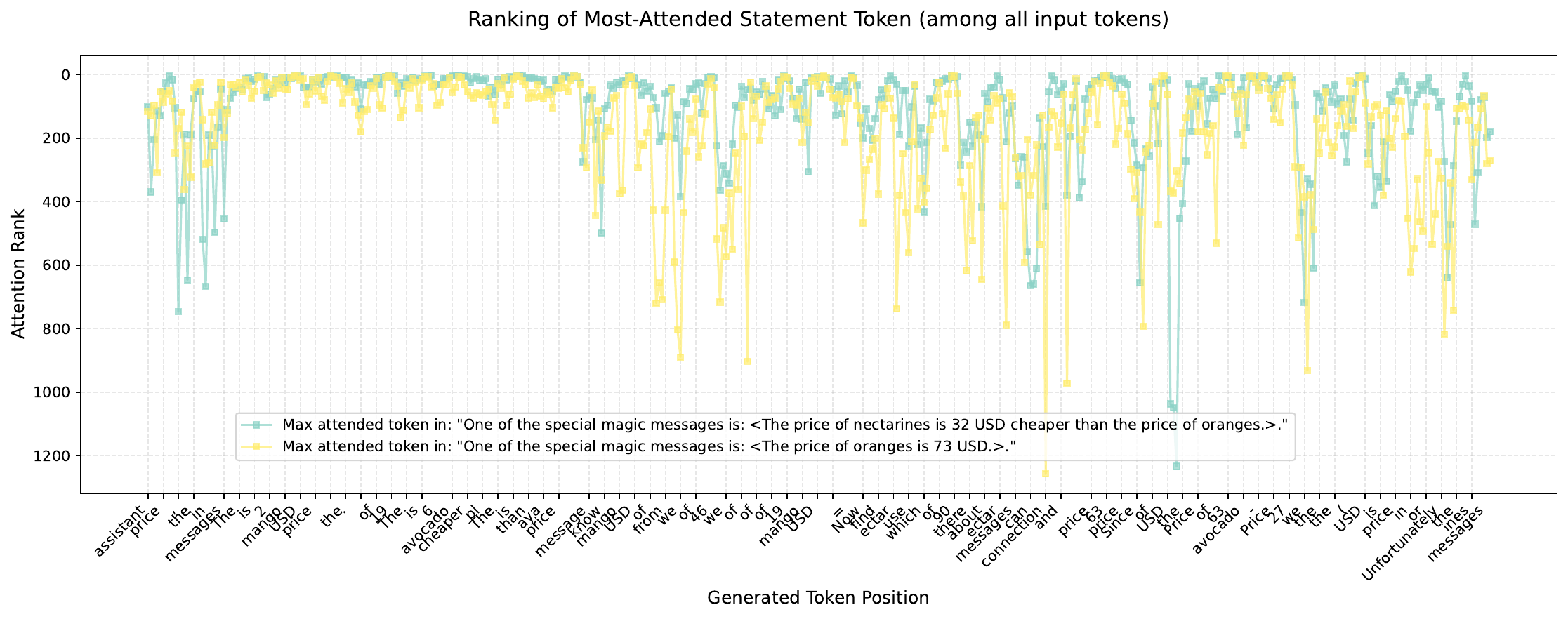}
    \caption{Ranking of tokens most attended in the statements. The example shows a failure case.}
    \label{fig:ranking}
\end{figure*}

\begin{figure*}
    \centering
    \includegraphics[width=0.95\textwidth, trim=0 0 0 20, clip]{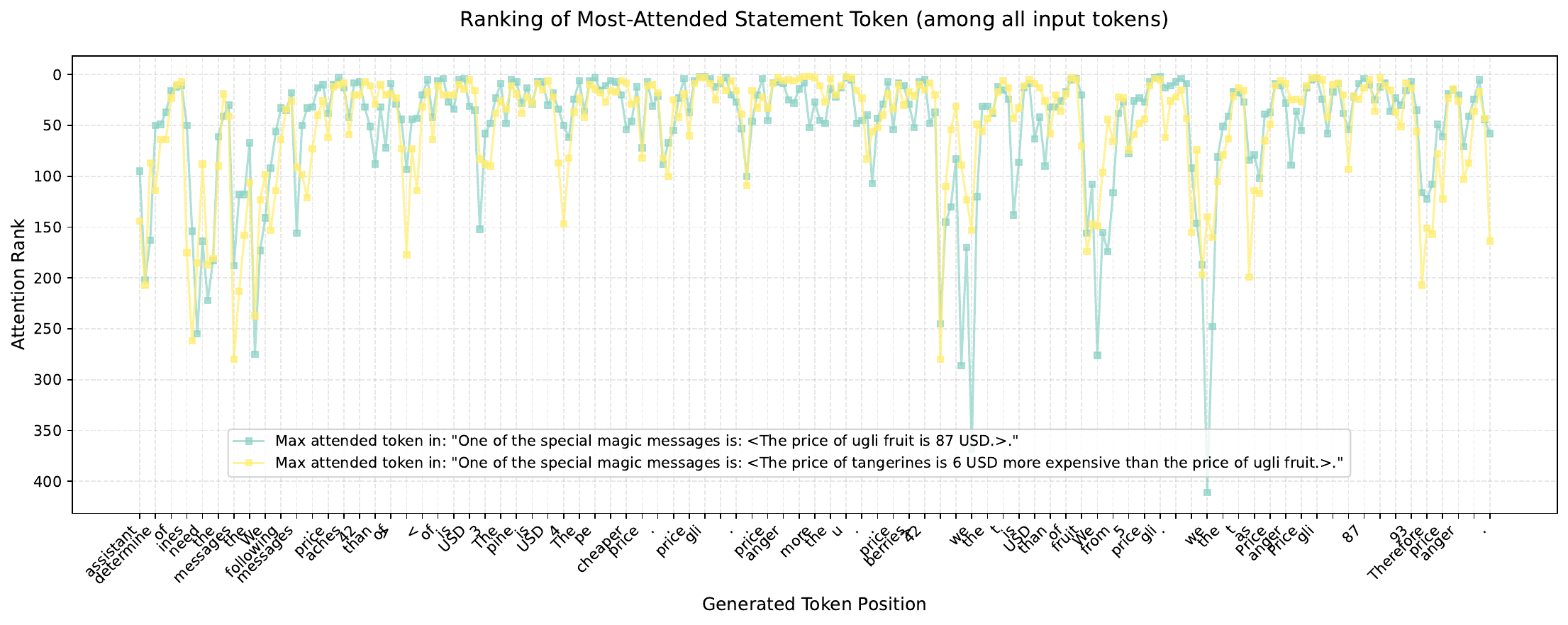}
    \caption{Ranking of tokens most attended in the statements. The example shows a success case.}
    \label{fig:ranking_success}
\end{figure*}

\begin{figure*}[t]
    \centering
    \includegraphics[width=0.95\linewidth, trim=0 20 0 0, clip]{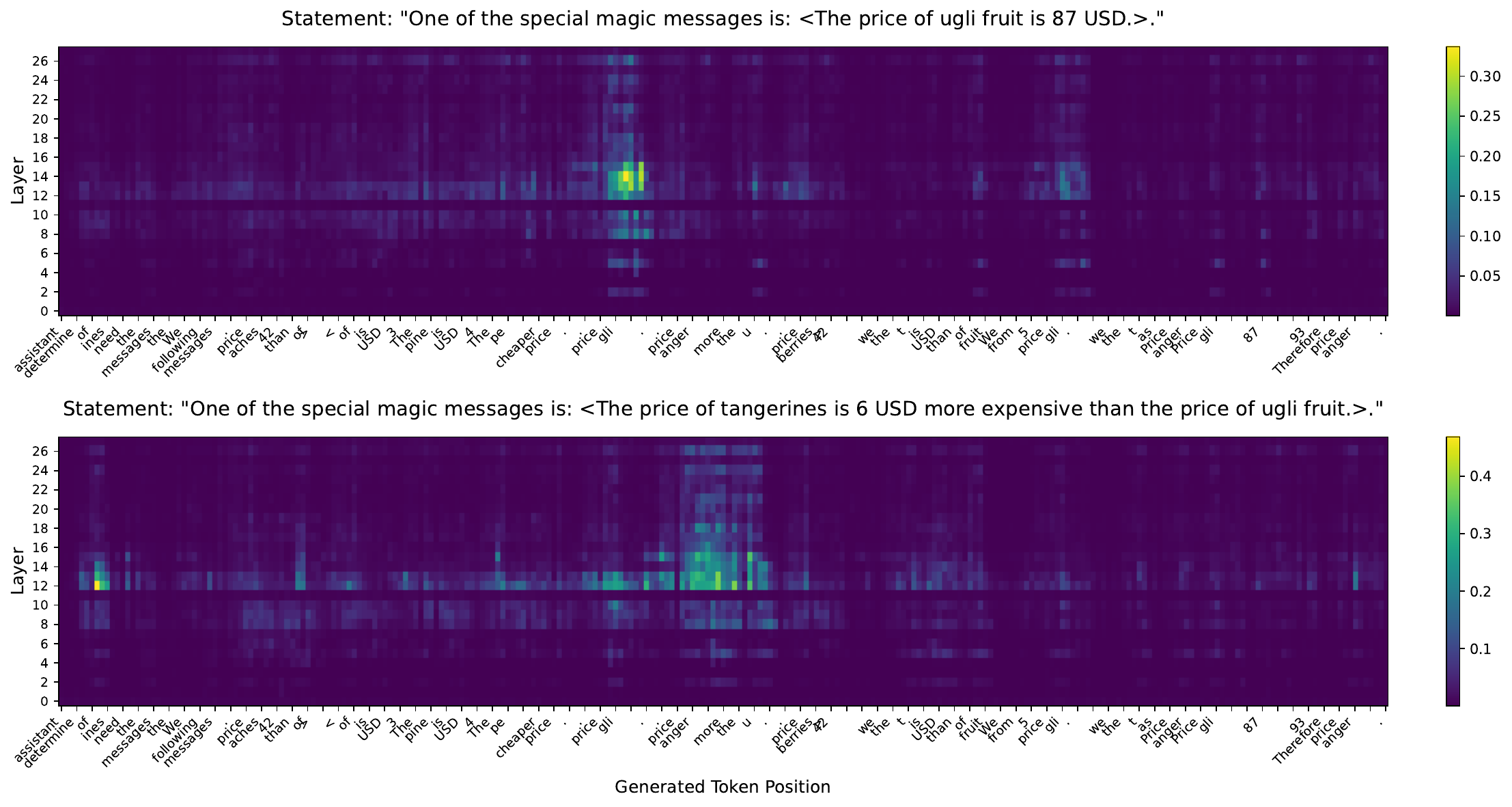}
    \vspace{-3mm}
    \caption{Proportion of attention from generated tokens to the input prompt across layers.}
    \label{fig:heatmap_success}
    \vspace{-3mm}
\end{figure*}

\section{Deduction: A Diagnostic Benchmark for Long-context Reasoning}
\label{sec:deduction}

We first create $6$ problem types including: fruit price, person age, car speed, city population, book length and planet temperature. For each of them, we generate $15$ entities as candidates. Then a statement is generated by first randomly sample a problem type and then a subset of entities is randomly sampled. Unique values are then assigned to these entities using a controlled random number generation process that ensures non-duplicative values. The statements further encodes relationships between entities by formulating independent and pairwise conditions based on templated statements, which describe direct values or comparative differences (e.g., “more expensive” or “older”). To increase the complexity and challenge of inference, additional distractor conditions involving extra entities are optionally introduced. Finally, a question is generated to prompt for numeric responses. Finally statements and distractors are randomly inserted to a haystack~\cite{LLMTest_NeedleInAHaystack} to flexibly extend the context length.

\section{Statement Ranking from the Attention}
\label{sec:statement_ranking}
In \autoref{sec:attention_retrieve}, we mention \autoref{fig:ranking} about the ranking of statement tokens. We now illustrate how we get the rankings. Specifically, we calculate the ranking by first identifying the most attended token within the statement 
$i^*(t)=\arg\max_{i\in I_{\text{stmt}}} A^{(l)}_{t,i},$
and then determine its rank among all the input tokens
\begin{equation}
    r(t)=1+\sum_{i=1}^{N} \mathbf{1}\Bigl( A^{(l)}_{t,i} > A^{(l)}_{t,i^*(t)} \Bigr).
\end{equation}
Apart from a failure case, we also show the heatmap plots and ranking plot for a successful retrieval case in \autoref{fig:heatmap_success} and \autoref{fig:ranking_success}.

\section{Case Study of Retrieved Facts}
\label{sec:case_study}
We show the cases where CoT suffers from implicit retrieval while {\ourmodel} can successfully retrieve the ground truth context.

\begin{table*}[h]
  \centering
  \begin{tabular}{|p{3cm}|p{4.5cm}|p{4.5cm}|}
    \hline
    \textbf{Ground Truth Facts or Answer} & \textbf{Original CoT} & \textbf{Retrieved Facts} \\
    \hline
    The length of Fables of Yldora is 40 pages shorter than the length of Saga of Lynora. The length of Saga of Lynora is 92 pages. & To answer this question, we need to refer to one of the special magic messages provided in the text. The message is:\texttt{\textbackslash n}\texttt{\textbackslash n}"One of the special magic messages is: <The length of Fables of Yldora is 40 pages.>."\texttt{\textbackslash n}\texttt{\textbackslash n}From this message, we can infer that the length of Fables of Yldora is 40 pages. & clause 136: One of the special magic messages is: <The length of Fables of Yldora is 40 pages shorter than the length of Saga of Lynora ... clause 460: One of the special magic messages is: <The length of Saga of Lynora is 92 pages.\\
    \hline
    Marie de' Medici & Based on the document, Philippe, Duke of Orléans (Document 53) was the younger son of Louis XIII of France and his wife, Anne of Austria. His mother, Anne of Austria, was the daughter of King Philip II of Spain and Elisabeth of Valois, making Elisabeth of Valois the grandmother of Philippe, Duke of Orléans. & clause 185:Document 54:Leonora Dori Leonora Dori Galiga (19 May 1568 – 8 July 1617) was a French courtier of Italian origin, an influential favourite of the French regent Marie de' Medici, mother of King Louis XIII of France ... clause 189:Document 56:Louise Élisabeth d'Orléans Louise Élisabeth d'Orléans was one of the seven daughters of Philippe II, Duke of Orléans, and his wife, Françoise Marie de Bourbon. \\
    \hline
    from 1986 to 2013 & The football manager who recruited David Beckham managed Manchester United between 1945 and 1969, and again for the second half of the 1970–71 season. The answer is: Sir Matt Busby managed Manchester United between 1945 and 1969, and again for the second half of the 1970–71 season. & clause 4:Document 2:\texttt{\textbackslash n}Alex Ferguson\texttt{\textbackslash n}Sir Alexander Chapman Ferguson, CBE (born 31 December 1941) is a Scottish former football manager and player who managed Manchester United from 1986 to 2013. \\
    \hline
  \end{tabular}
  \caption{Cases where CoT is unable to retrieve the ground truth facts but {\ourmodel} can successfully retrieve.}
\end{table*}

\section{Prompt Used to Integrate Facts}\label{sec:prompt}
We use the same prompt for Deduction and two QA datasets. The template is as follows:
\begin{adjustwidth}{-5mm}{-5mm}
    \begin{quotation}
    \noindent \textit{``\{anything before question\}\\\\Some clauses extracted from the context that might be related:\\\{clauses\}\\\\\{anything after question starts\}''}
    \end{quotation}
\end{adjustwidth}
For BABILONG, we employ a slightly different template since these tasks are sensitive fact order.
\begin{adjustwidth}{-5mm}{-5mm}
    \begin{quotation}
    \noindent \textit{``\{anything before question\}\\\\Some clauses are extracted from the context that might be related:\\\{clauses\}\\\\Notice that the clause indices represents the order of them appearing in the context. Larger clause indices indicate that they appear later in the context. The answer to the question is sensitive to the order in the context. The clauses only serve as a hint, please check the original context for exact information.\\\\\{anything after question starts\}''}
    \end{quotation}
\end{adjustwidth}
We reorder the retrieved facts according to the order they appear in the context.

\end{document}